\documentclass[runningheads]{llncs}
\usepackage[T1]{fontenc}
\usepackage{graphicx}
\usepackage{CJKutf8}
\usepackage{enumitem}
\usepackage{amsmath}
\usepackage{amssymb}
\usepackage{amsfonts}
\usepackage{xcolor}
\usepackage{graphicx}
\usepackage{appendix}
\usepackage{ulem}

\usepackage{tabu}
\usepackage{multicol}
\usepackage{multirow}
\usepackage{float}
\usepackage{makecell}
\usepackage{booktabs}

\usepackage[misc]{ifsym}

\begin{document}

\title{In-Context Learning for Knowledge Base Question Answering for Unmanned Systems based on Large Language Models}
\titlerunning{In-Context Learning for KBQA based on Large Language Models}
%
\author{
    Yunlong Chen\inst{1} \and
    Yaming Zhang\inst{1} \and
    Jianfei Yu\inst{1}\textsuperscript{(\Letter)} \and
    Li Yang\inst{2} \and
    Rui Xia\inst{1}
}
\authorrunning{Y. Chen et al.}
%
\institute{
    School of Computer Science and Engineering, \\Nanjing University of Science and Technology, China \and
    Wee Kim Wee School of Communication and Information, \\Nanyang Technological University, Singapore\\
    \email{\{ylchen, ymzhang, jfyu, rxia\}@njust.edu.cn}
}
\maketitle              

\begin{abstract}
Knowledge Base Question Answering (KBQA) aims to answer factoid questions based on knowledge bases. 
However, generating the most appropriate knowledge base query code based on Natural Language Questions (NLQ) poses a significant challenge in KBQA.
In this work, we focus on the CCKS2023 Competition of Question Answering with Knowledge Graph Inference for Unmanned Systems.
Inspired by the recent success of large language models (LLMs) like ChatGPT and GPT-3 in many QA tasks, we propose a ChatGPT-based Cypher Query Language (CQL) generation framework to generate the most appropriate CQL based on the given NLQ.
Our generative framework contains six parts: 
an auxiliary model predicting the syntax-related information of CQL based on the given NLQ, 
a proper noun matcher extracting proper nouns from the given NLQ, 
a demonstration example selector retrieving similar examples of the input sample,
a prompt constructor designing the input template of ChatGPT, 
a ChatGPT-based generation model generating the CQL, 
and an ensemble model to obtain the final answers from diversified outputs.
With our ChatGPT-based CQL generation framework, we achieved the second place in the CCKS 2023 Question Answering with Knowledge Graph Inference for Unmanned Systems competition, achieving an F1-score of 0.92676.

\end{abstract}

\begin{CJK*}{UTF8}{gbsn}

\section{Introduction}

\begin{figure}[!tp]
\centering
\includegraphics[scale=0.5]{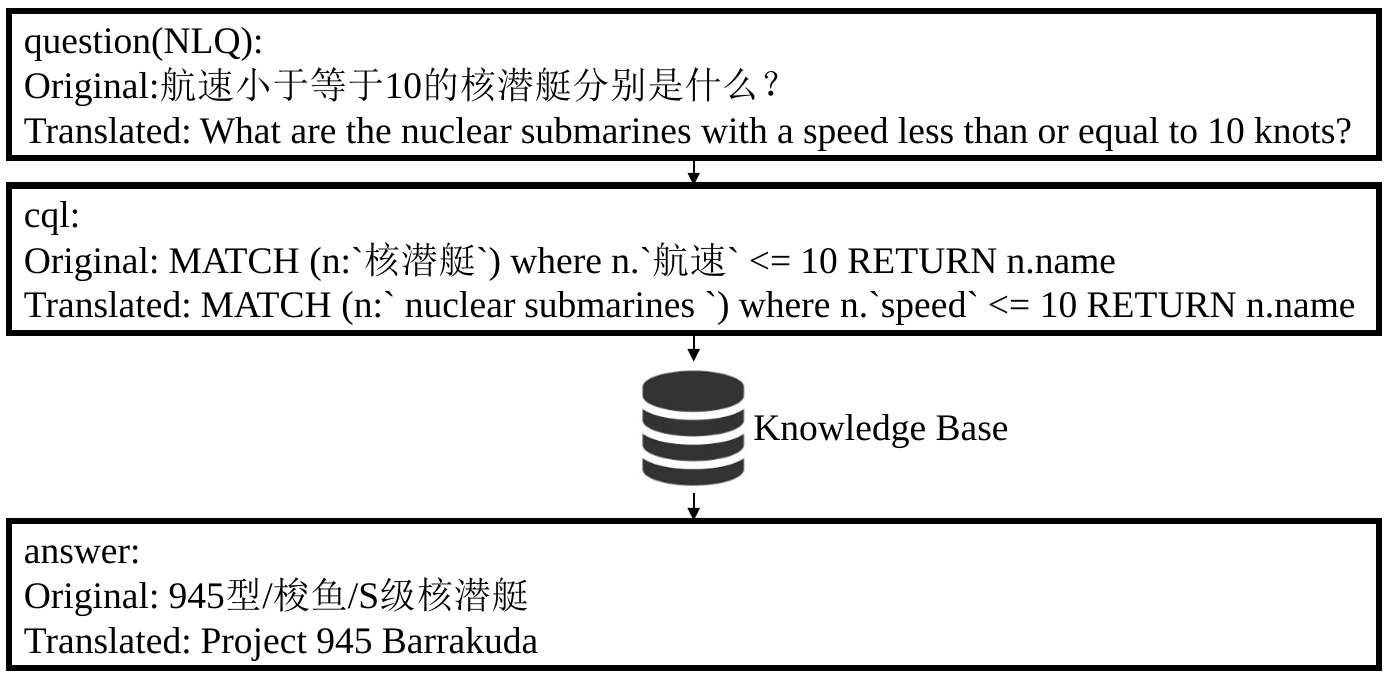}
\caption{\label{dataset_example}Illustration of an example in the evaluation task.}
\end{figure}

As an important task in Natural Language Processing (NLP), Knowledge Base Question Answering (KBQA) aims to generate accurate and complete query statements from user-provided natural language questions (NLQs), and these query statements are then used to retrieve relevant information from the knowledge base and provide accurate answers.
In this work, we focus on the CCKS 2023 Question Answering with Knowledge Graph Inference for Unmanned Systems competition, which is a KBQA evaluation task where cypher query language (CQL) serves as the query statements.
Fig. \ref{dataset_example} gives an example of the CCKS2023 competition.

In the literature, most existing studies on KBQA can be categorized into two types: information retrieval-based (IR-based) approaches and semantic parsing-based (SP-based) approaches.
Both of them require first identifying the subject within the NLQ and linking it to an entity in the knowledge base (KB).
The former line of work aims to derive answers by reasoning within a question-specific graph extracted from the KB with the assistance of those linked entities \cite{yan2021large,zhang2022subgraph}, whereas the latter line of work aims to obtain answers by executing a parsed logic form based on the linked entities \cite{gu2022knowledge,gu2021beyond}.
Since the annotation in the dataset of the CCKS2023 competition contains manually annotated CQLs, we follow the latter line of approaches in this work.

However, the majority of existing SP-based approaches are built upon LSTM or pre-trained models like BERT, which are constrained by their scale or their pre-training data and may encounter challenges in effectively generating suitable knowledge base query codes based on NLQs.
With the recent advancements of pre-trained language models, many Large Language Models (LLMs) have been shown to achieve surprisingly good performance on many question answering datasets under zero-shot or few-shot settings. 
These LLMs have also showcased an impressive capacity to deeply comprehend sentence semantics and accurately translate them into multiple languages, and even generate code when required. 
Therefore, we aim to explore the potential of LLMs on the Chinese KBQA task for unmanned systems.

Specifically, we propose a ChatGPT-based CQL generation framework, consisting of six parts. 
The first part involves an auxiliary model that takes the given NLQ as input and predicts structural information for each clause separately. 
The second part comprises a proper noun matcher, which identifies explicitly mentioned proper nouns existing in the KB from the given NLQ. 
The third part consists of a demonstration example selector that employs a key-information-based similarity calculation criterion to retrieve demonstration samples for the given NLQ based on the aforementioned results.
The fourth part encompasses a prompt constructor that constructs input text by integrating demonstration samples, NLQ, and task-specific prior knowledge. 
The fifth part incorporates a ChatGPT-based generation model, which inputs the constructed text into ChatGPT to generate CQL. Subsequently, post-processing is applied to the generated CQL. 
Lastly, the sixth part introduces an ensemble model, in which multiple answers retrieved from the knowledge base by the post-processed CQL are combined through a voting mechanism to obtain the final result.

We conduct experiments on the dataset provided by the competition, and the results show the high efficiency of our generative framework. 
Therefore, we achieved the second place in the CCKS 2023 Question Answering with Knowledge Graph Inference for Unmanned Systems competition with an F1-score of 0.92676.

\section{Related Work}

\subsection{Large Language Model}
\label{LLM}
Large Language Models (LLMs) typically possess a vast number of learnable parameters and undergo extensive training on enormous text datasets, examples of which include ChatGPT\cite{openai-chatgpt}, LLaMA\cite{touvron2023llama}, OPT\cite{zhang2022opt}, PaLM\cite{chowdhery2022palm}, CodeX\cite{chen2021evaluating}, 
and so on. 
With the advancement of LLMs, traditional pre-trained models like BERT\cite{devlin2018bert}, RoBERTa\cite{liu2019roberta}, BART\cite{lewis2019bart}, 
T5\cite{raffel2020exploring}, have faced great challenges. 
The ability of LLMs to adapt to downstream tasks without the need for retraining, but task-specific instructions, has greatly reduced the cost of solving downstream tasks.

\subsection{In-Context Learning}
As mentioned in Section \ref{LLM}, LLMs typically demonstrate emergent abilities \cite{wei2022emergent,brown2020language} with increasing model and corpus size, i.e., the ability to learn from the given examples present in the context, known as In-Context Learning (ICL). 
This ability helps LLMs in better adapting to downstream tasks. While solely relying on task-specific instructions may not lead to superior performance compared to fine-tuned models in some downstream tasks, introducing ICL can often result in considerable improvements in LLMs' performance on downstream tasks.

\subsection{Chain-Of-Thought}
Chain-of-Thought (CoT) \cite{wei2022chain} is an extremely efficient and easy prompting strategy that endows LLMs with reasoning capabilities, enabling LLMs to decompose and comprehend complex tasks. 
Specifically, CoT leverages several given examples with inferred answers to assist LLMs in comprehending the reasoning process of complex tasks, thus performing reasoning on the target problem and obtaining results.
In general, CoT leverages several pre-given exemplars with inferred answers to help LLMs understand the reasoning process of intricate tasks.


\section{Methodology}

Recently, LLMs have showcased robust generalizability across a diverse spectrum of tasks by leveraging few-shot in-context learning. Notably, LLMs possess the capability to transform unstructured sentences into structured and executable code, rendering them valuable assets in KBQA \cite{li2023few}.

However, the CQL solely generated by ChatGPT falls short of our expectations. 
Therefore, we observe CQL's overall structure and summarize empirical knowledge to design processing techniques and auxiliary tasks. 
These aid ChatGPT in capturing key information and parsing CQL structures from NLQs, enabling it to adapt to downstream tasks and generate high-quality CQL.

\begin{figure}[!tp]
\centering
\includegraphics[scale=0.35]{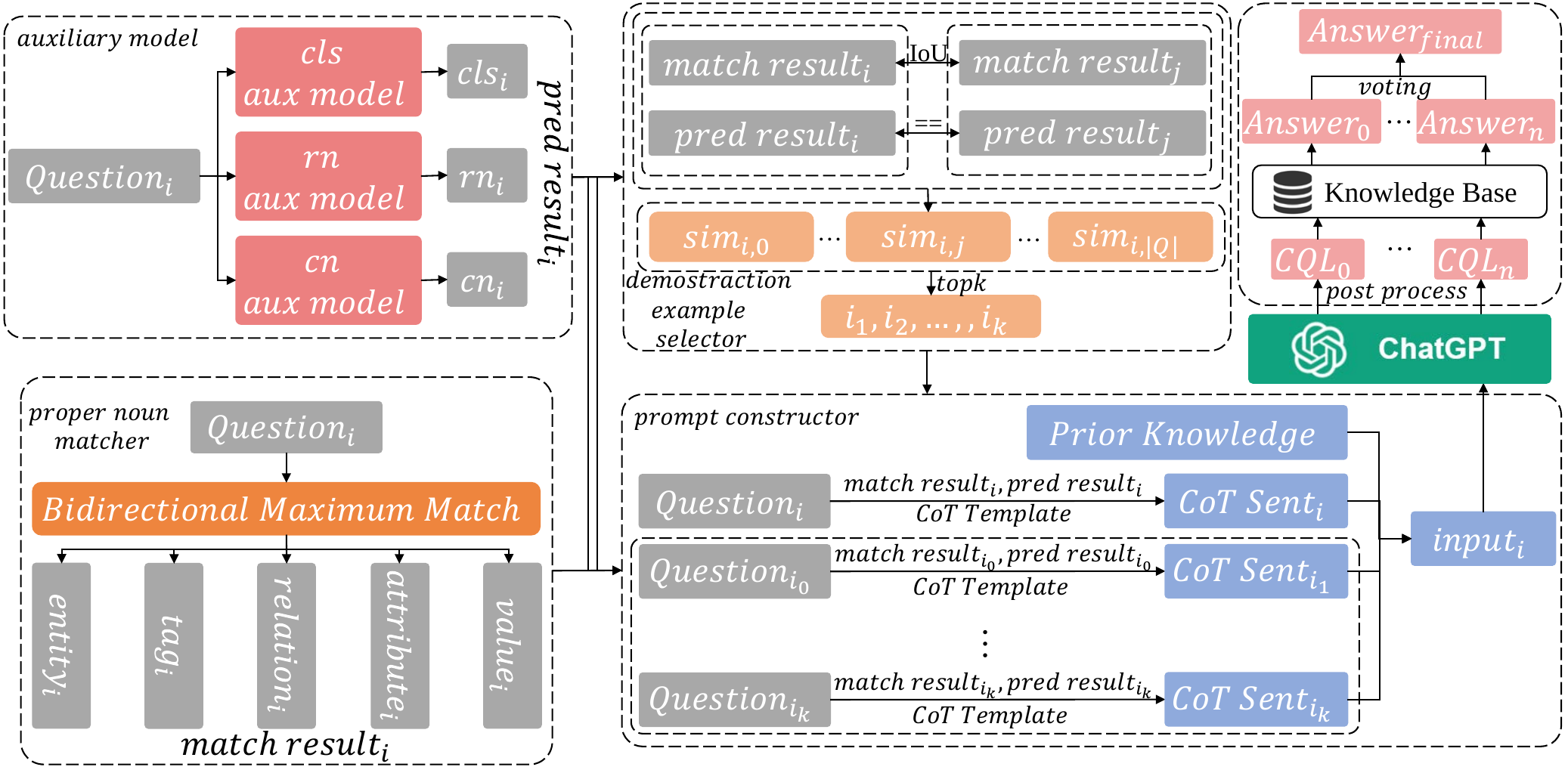}
\caption{\label{framework}The overall architecture of our ChatGPT-based KBQA framework.}
\end{figure}

\begin{figure}[!tp]
\centering
\includegraphics[scale=0.4]{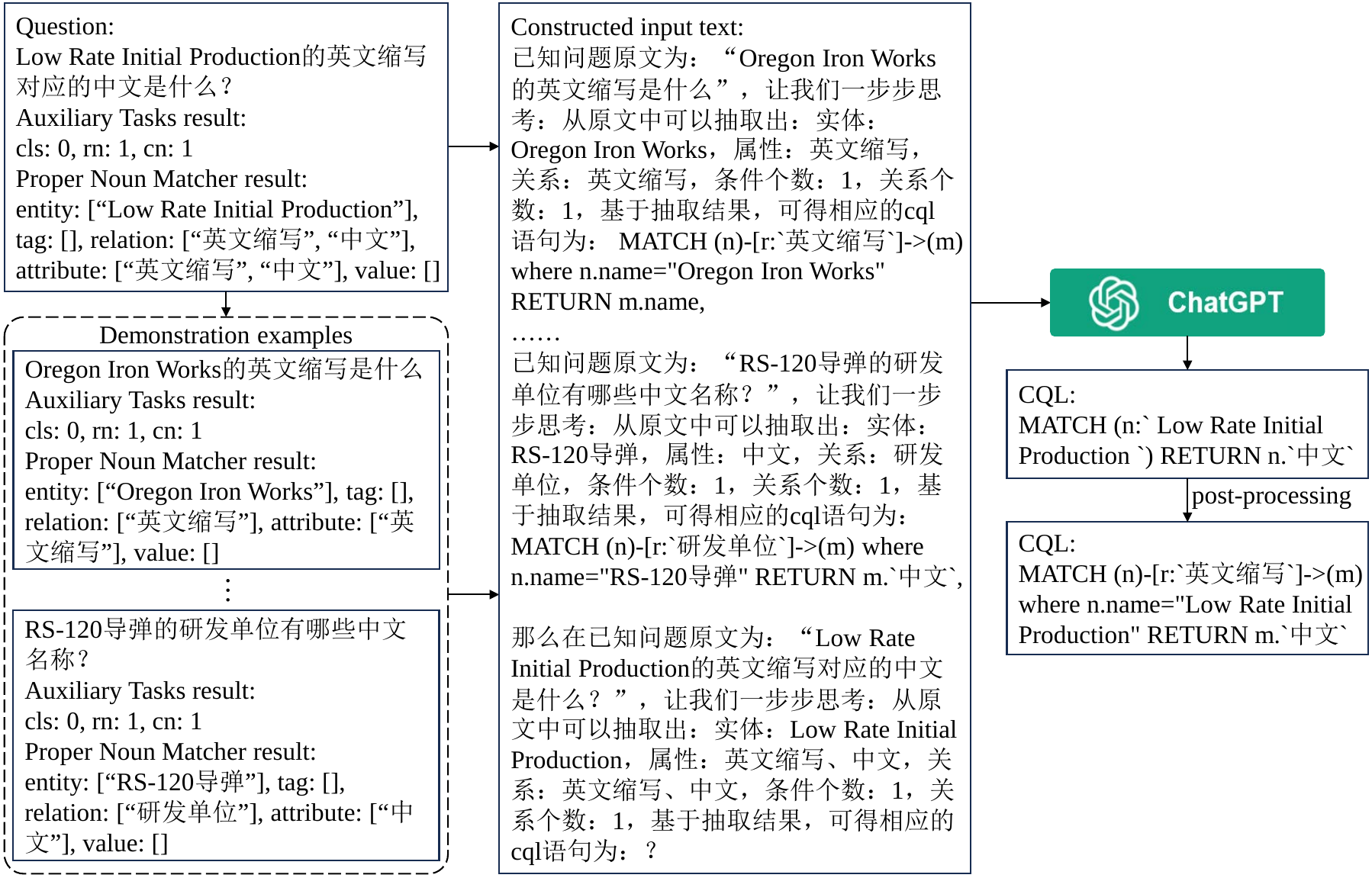}
\caption{\label{pipeline}A visualized example of generating CQL from NLQ.}
\end{figure}

In general, our ChatGPT-based CQL generation framework consists of six steps (excluding KB construction and answer retrieval), as illustrated in Fig.~\ref{framework} and Fig.~\ref{pipeline} provides a visualized example of using our generative framework to generate CQL from NLQ.

\begin{enumerate}[label=\arabic*.]
    \item Three auxiliary tasks to predict structural information for CQL clauses.
    \item Bidirectional maximum matching-based proper noun matching for NLQ.
    \item Selecting demonstration examples based on the aforementioned results.
    \item Combining NLQ, demonstration examples, and prior knowledge into CoT format as input text.
    \item ChatGPT generates CQL based on the constructed input text, followed by post-processing of the generated CQL.
    \item Voting for the answers retrieved from the given KB by CQLs.
\end{enumerate}

The overall workflow is as follow:

Firstly, within the \textbf{auxiliary model}, the structural information of different clauses in CQL ($pred~result_i$) is predicted based on NLQ ($Question_i$). 
Meanwhile, the \textbf{proper noun matcher} identifies the proper nouns ($match~result_i$) explicitly mentioned in $Question_i$ and existing in the KB.

Subsequently, both $pred~result_i$ and $match~result_i$ are fed into the \textbf{demonstration example selector} to compute the similarity between different samples, thereby selecting demonstration examples ($i_1, i_2 \dots, i_k$) for each sample.

Following that, within the \textbf{prompt constructor}, the $Question_i$, $pred~result_i$, and $match~result_i$ are firstly combined into CoT format to obtain $CoT~Sent_i$. 
Next, $Prior~Knowledge, CoT~Sent_i$, and corresponding demonstration examples ($CoT~Sent_{i_1}, \dots, CoT~Sent_{i_k}$) are combined to form $input_i$. 
Then, $input_i$ is fed into \textbf{ChatGPT} to generate CQL, followed by post-processing.

By repeating the aforementioned steps, multiple CQLs ($CQL_0, \dots, CQL_n$) are acquired. Executing these CQLs in the given KB yields multiple answers ($Answer_0, \dots, Answer_n$). 
Employing a \textbf{voting} mechanism on these answers yields the most reliable response as the final outcome ($Answer_{final}$).

\subsection{Auxiliary Tasks}
\label{aux_task_section}

Upon analyzing CQLs and evaluating the task instructions, we summarize four execution intent categorized by the content of the \textsf{RETURN} clause: 1. Entity-Attribute retrieval; 2. Entity counting; 3. Conditional sorting; 4. Attribute-value comparison.
These diverse execution purposes impact the overall CQL structure. Furthermore, the CQL structure is influenced by the number of relations and conditions, referring to the number of entity jumps in the \textsf{MATCH} clause and the restrictive conditions in the \textsf{WHERE} clause.

Due to the potential impact of the aforementioned information on the CQL structure and the difficulty in directly obtaining them from NLQs or other related sources, we design three auxiliary tasks for corresponding predictions:

\renewcommand{\labelitemi}{\textbullet}
\begin{itemize}
    \item Intent classification ($cls$).
    \item Relation count classification ($rn$).
    \item Condition count classification ($cn$).
\end{itemize}

The aforementioned auxiliary tasks take NLQs as input and predict the intent (0/1/2/3), relation number (0/1/2), and condition number (0/1/2) of the CQL.
For example, given the CQL in Fig.1, the output of the three auxiliary tasks are 0, 0, and 1, respectively. These values signify that the intent of the CQL is to retrieve an entity's attribute, there are no relations mentioned in the CQL, and there is one condition specified.

Notably, these three auxiliary tasks share NLQs as input and yield similar outputs, all treated as simple classification results, thus allowing them to utilize a consistent model architecture.
Any pre-trained model like BERT-chinese and mT5 can be used to extract global features from the text, followed by a feed-forward network for output prediction of the auxiliary tasks.

\subsection{Proper Noun Matcher}
\label{match_section}

In NLQs, some proper nouns are either explicitly or implicitly mentioned, which are likely to appear in the CQLs. 
Therefore, it is essential to extract the mentioned proper nouns from NLQs. 
To this end, we employ the bidirectional maximum matching based on the given proper noun vocabulary to all NLQs, obtaining the proper nouns that appear in them.

During bidirectional maximum matching, we observed that unconstrained execution could introduce noise into the matching process for relation and attribute proper nouns. 
Although designing a universal set of constraints is challenging, creating NLQ-specific constraints is more feasible.
Therefore, our approach gives precedence to extracting entities and tags from the NLQ. 
Subsequently, by leveraging relevant knowledge from KB, we integrate the entities corresponding to the tags into the matched entities.
Taking the entities as the starting point and considering two-hop relations as the scope, all the involved entities are referred to as entity set, and all the involved relations are referred to as \textit{candidate} relation set.
Using the entity set, we retrieve associated attributes from the given knowledge base, creating a \textit{candidate} attribute set.

The NLQ's bidirectional maximum matching results in the \textit{matched} attribute and relation sets.
The final matching results for attribute and relation sets are obtained by intersecting the \textit{candidate} sets with the \textit{matched} sets.

\subsection{Demonstration Example Selector}
\label{similarity_section}

For NLQ-specific demonstration example selection, an appropriate similarity calculation criterion is vital. 
The conventional method, such as using BERT or similar models for cosine similarity computation between global features (GF-Sim), may not yield accurate similarity results due to substantial noise present in NLQs. This noise considerably affects the results of GF-Sim. 
With approximately 35\% of the text containing valid information (proper nouns) while the rest being noisy, the impact on GF-Sim is significant.

Based on the above findings, we propose a key-information-based similarity criterion (KI-Sim) for NLQs. It focuses on key components in NLQ like proper nouns and other influential details impacting CQLs, which is computed as follows: 
\begin{align}
    \label{similarity_equation}
    Similarity(i, j) &= \sum_k^{e, t, r, a, v} w_k * \text{IoU}(i_k, j_k) + \sum_k^{cls, rn, cn} w_k * (i_k == j_k)
\end{align}
where, $i, j$ represent NLQs' id, 
$e, t, r, a, v$ stand for entity, tag, relation, attribute, and value, 
$cls, rn, cn$ indicate the predicted intent, relation number, and condition number, 
$i_k, j_k$ refer to proper nouns or auxiliary task predictions from NLQs with id $i, j$, 
and $w_k$ signifies the corresponding similarity weight.

\subsection{Prompt Constructor}

\subsubsection{Prior Knowledge}

To bolster ChatGPT's grasp and alignment with the downstream task, we propose to incorporate task-specific prior knowledge and integrate it into text (see Appendix 1) to feed into ChatGPT. 
These prior knowledge are derived from observations of CQLs and hold general applicability for this downstream task, rather than being specific to any particular NLQ.

\subsubsection{In-Context Learning}

To enhance ChatGPT's CQL generation, we utilize KI-Sim (Section \ref{similarity_section}) to select demonstration examples for ICL. This aids ChatGPT in CQL generation, thereby improving the quality of the generated CQLs.

\subsubsection{Chain-Of-Thought}

Intuitively, directly generating CQL from NLQ is challenging. Yet, analyzing the composition and syntax of CQLs and NLQs reveals a high likelihood of shared proper nouns. Drawing inspiration from the CoT method, we split the CQL generation task into two sub-tasks:

\begin{enumerate}[label=\arabic*.]
    \item Matching relevant proper nouns from NLQ.
    \item Generating CQL based on NLQ and the matched proper nouns.
\end{enumerate}

As elucidated in Section \ref{match_section}, with the completion of sub-task 1, ChatGPT now focuses on addressing sub-task 2. To achieve this, a template was devised (see Appendix 2) that integrates matched proper nouns and NLQ following the CoT approach, presenting them jointly to ChatGPT.

\subsection{ChatGPT-Based Generation Model}
\label{post_process_section}

After inputting the constructed text into ChatGPT, it generates the corresponding CQL for the given NLQ. 
Concerning the CQL, three common situations arise:

\renewcommand{\labelitemi}{\textbullet}
\begin{itemize}
    \item Entities matched are mis-classified as tags.
    \item The number of relation does not match the results from auxiliary tasks.
    \item Matching with unprovided proper nouns.
\end{itemize}

Among these situations, the first two can result in inaccurate CQL execution and should be prevented or rectified. In contrast, the third situation is favorable. As described in Section \ref{match_section}, only explicit proper nouns can be matched, leaving implicit ones unmatched. This implies ChatGPT's successful identification of implicit proper nouns in the NLQ. In this case, post-processing methods can be used to map non-proper nouns to the provided vocabulary for correction.

Consequently, we present three post-processing methods to address these situations to get the final ChatGPT-generated CQL:

\renewcommand{\labelitemi}{\textbullet}
\begin{itemize}
    \item Reclassify the mis-classified tags as entities and position them correctly.
    \item Use condition truncation and filling for isolated condition clause correction.
    \item Utilize fuzzy matching to map implicit proper nouns to vocabulary and make CQL modifications correspondingly.
\end{itemize}

\subsection{Ensemble Model}

Executing CQL in the KB doesn't always ensure correct answers, and sometimes no answers are found. 
However, generating new CQLs could enhance the retrieval success. 
Thus, for each NLQ, we generate multiple CQLs, retrieve corresponding answers, and apply a voting mechanism to ascertain the final answer.

\section{Experiment}

\subsection{Dataset}

We conduct experiments based on the competition's dataset, which encompasses knowledge base construction data, as well as training, validation (preliminary round), and test (final round) datasets. 
The training set includes annotations, while validation's annotations are released with the un-annotated test set in the final round. 
Annotations include answers and the CQL used for retrieval from KB. We will validate our generative framework on this dataset.

\subsection{Evaluation}

In the experiments, the main evaluation concern is precise answer retrieval for NLQs.
The evaluation metrics include Macro Precision (Eqn.~(\ref{evaluation_metric_p})), Macro Recall (Eqn.~(\ref{evaluation_metric_r})), and Averaged F1 (Eqn.~(\ref{evaluation_metric_f1})), which are defined below: 
\begin{align}
    \label{evaluation_metric_p}
    &P = \frac{1}{|Q|} \sum_{i=1}^{|Q|} P_i, ~~~P_i = \frac{|A_i \cap G_i|}{|A_i|} \\
    \label{evaluation_metric_r}
    &R = \frac{1}{|Q|} \sum_{i=1}^{|Q|} R_i, ~~~R_i = \frac{|A_i \cap G_i|}{|G_i|} \\
    \label{evaluation_metric_f1}
    &F1 = \frac{1}{|Q|} \sum_{i=1}^{|Q|} \frac{2 P_i R_i}{P_i + R_i}
\end{align}
where $|Q|$ denotes the number of NLQs in the dataset, $A_i, G_i$ denotes the player's and ground-truth answer sets to the question whose id is $i$, respectively.

\subsection{Implementation}

\subsubsection{Similarity}
During similarity computation, entity weights are set to 5, tag weights to 3, relation weights to 3, attribute weights to 1, value weights to 0.5, $cls$ weights to 0.5, $rn$ weights to 0.3, and $cn$ weights to 0.3.

\subsubsection{ChatGPT}
The ChatGPT we used in this paper is \textsf{gpt-3.5-turbo-0613}. 
It should be noted that we set the temperature parameter to 1 (default) to ensure the diversity of CQLs when ChatGPT generates responses multiple times.

\subsubsection{Auxiliary Task}
\label{aux_task_setting}
We use the \textsf{mT5-large} as the pre-trained model. 
For the auxiliary tasks, the global random seed is 33. 
The batch size is 32, trained for 100 epochs. 
Initial learning rates for the backbone and non-backbone part are set at 1e-6 and 1e-4, respectively. 
Cross-entropy loss is employed for loss calculation.

\subsection{Main Results}

\begin{table}
    \caption{Main Results. Note the Prior indicate the prior knowledge, the Ensemble indicate the ensemble model, the Post indicate the post-processing in ChatGPT-based Generation Model.}
    \label{main_result}
    \centering
    \renewcommand{\arraystretch}{1.35}
    \resizebox{\linewidth}{!}
    {
        \begin{tabular}{p{1.5cm}p{2cm}p{2cm}p{1.5cm}p{2cm}p{2cm}}
            \Xhline{2pt}
            \makecell[c]{+Prior} & \makecell[c]{+Ensemble} & \makecell[c]{+ICL+CoT} & \makecell[c]{+Post} & \makecell[c]{Averaged F1\\(Validation)} & \makecell[c]{Averaged F1\\(Test)}\\
            \Xhline{1pt}
            \makecell[c]{\checkmark} & \makecell[c]{$\times$}& \makecell[c]{$\times$}& \makecell[c]{$\times$}& \makecell[c]{0.72539} & \makecell[c]{\textbackslash} \\
            \makecell[c]{\checkmark} & \makecell[c]{\checkmark} & \makecell[c]{$\times$}& \makecell[c]{$\times$}& \makecell[c]{0.83865} & \makecell[c]{0.86204} \\
            \makecell[c]{\checkmark} & \makecell[c]{\checkmark} & \makecell[c]{\checkmark} & \makecell[c]{$\times$}& \makecell[c]{\textbackslash} & \makecell[c]{0.91561} \\
            \makecell[c]{\checkmark} & \makecell[c]{\checkmark} & \makecell[c]{\checkmark} & \makecell[c]{\checkmark} & \makecell[c]{\textbackslash} & \makecell[c]{\textbf{0.92676}} \\
            \Xhline{2pt}
        \end{tabular}
    }
\end{table}

In Table \ref{main_result}, the performance of ChatGPT with different processing techniques is presented, where the last row shows the performance of our proposed ChatGPT-based CQL generation framework. 

In the preliminary round, with only prior knowledge and voting mechanism, our F1 score on the validation set is 0.83865, obtaining the second place. 
In the final round, our ChatGPT-based CQL generation framework achieves an F1 score of 0.92676 on the test set, obtaining the second place.

\subsection{Ablation Study}
As shown in Table \ref{main_result}, all different processing techniques can improve the final performance, but their effects are different:

\subsubsection{Ensemble Model}
The essence of this technique is to allow ChatGPT to generate multiple CQLs and vote on the answers. 
The multiple generations can help ChatGPT re-understand NLQ and increase the diversity of generated CQLs.

\subsubsection{ICL+CoT}
The essence of this technique is to enable ChatGPT to capture and learn implicit relations that may exist in downstream tasks based on given demonstration examples. 
By using the decomposed sub-tasks, ChatGPT can achieve a deeper understanding of the downstream task and adapt to it, generating higher quality and more robust CQLs.

\subsubsection{Post-Processing}
The essence of this technique is to manually correct the generation errors of ChatGPT without interfering with its process of generating CQLs. Instead, it intervenes in the results generated by ChatGPT, ensuring that the results do not contain factual errors.

\subsection{Auxiliary Task Results}

\begin{table}
    \caption{The performance on three auxiliary tasks}
    \label{aux_task_result}
    \centering
    \renewcommand{\arraystretch}{1.35}
    \setlength{\tabcolsep}{4mm}
    {
        {
            \begin{tabular}{cc}
                \Xhline{2pt}
                \makecell[c]{Auxiliary Tasks} & \makecell[c]{Accuracy(\%)} \\
                \Xhline{1pt}
                \makecell[c]{Intent classification} & 99.0 \\
                \makecell[c]{Relation count classification} & 97.0 \\
                \makecell[c]{Condition count classification} & 98.2 \\
                \Xhline{2pt}
            \end{tabular}
        }
    }
\end{table}

Based on the performance of the auxiliary tasks in Table \ref{aux_task_result}, we can find that the proposed model performs well on the three auxiliary tasks. 
Therefore, it is generally useful to incorporate the auxiliary task-related information into our generative framework.

\subsection{Similarity Comparison}
\label{similarity_compare_section}

To verify KI-Sim's effectiveness (Section \ref{similarity_section}), we present the demonstration examples in Table \ref{sim_example_comparison}. 
The global features are extracted from \textsf{bert-base-chinese}.

\begin{table*}
    \centering
    \caption{Demonstration examples based on different similarity calculation criterion}
    \label{sim_example_comparison}
    \renewcommand{\arraystretch}{1}
    \resizebox{\linewidth}{!}{
        \begin{tabular}{cc|c|c}
            \Xhline{2pt}

            \multicolumn{2}{c|}{\multirow{2}{*}{NLQ}} & \makecell[c]{Original} & \makecell[c]{Translated} \\ \cline{3-4}
            & & \makecell[c]{\textbf{\fcolorbox{green}{white}{最大飞行速度}}小于等于\fcolorbox{green}{white}{\textbf{460}}的实体有几个？} & \makecell[c]{How many entities have a \fcolorbox{green}{white}{\textbf{maximum}}\\ \fcolorbox{green}{white}{\textbf{flying speed}} less than or equal to \fcolorbox{green}{white}{\textbf{460}}?} \\ [2ex]

            \Xhline{1pt}

            \multicolumn{1}{c}{\multirow{3}{*}{\makecell[c]{GF-\\Sim}}} & \makecell[c]{top1} & \makecell[c]{\textbf{\fcolorbox{red}{white}{阿姆德-500M/2M沉底水雷}}的\textbf{\fcolorbox{red}{white}{产国}}是哪个？} & \makecell{What is the {\fcolorbox{red}{white}{\textbf{origin country}}} of \\the {\fcolorbox{red}{white}{\textbf{AMD-500M/2M Submarine Mine}}}?}\\

            & \makecell[c]{top2} & \makecell[c]{\textbf{\fcolorbox{red}{white}{94式90毫米轻迫击炮}}的\textbf{\fcolorbox{red}{white}{口径}}是多少？} & \makecell[c]{What is the {\fcolorbox{red}{white}{\textbf{caliber}}} of the \\{\fcolorbox{red}{white}{\textbf{Type-94 90mm Light Mortar}}}?} \\
     
            & \makecell[c]{top3} & \makecell[c]{\textbf{\fcolorbox{red}{white}{弹径}}为\textbf{\fcolorbox{red}{white}{1.37}}的\textbf{\fcolorbox{red}{white}{舰地（潜地）导弹}}有哪些？} & \makecell[c]{Which \fcolorbox{red}{white}{\textbf{Ship-to-Ground (Submarine-to-}}\\ \fcolorbox{red}{white}{\textbf{Ground) Missile}} has a {\fcolorbox{red}{white}{\textbf{caliber}}} equal to {\fcolorbox{red}{white}{\textbf{1.37}}}?} \\ [2ex]

            \Xhline{1pt}

            \multicolumn{1}{c}{\multirow{3}{*}{\makecell[c]{KI-\\Sim}}} & \makecell[c]{top1} & \makecell[c]{\textbf{\fcolorbox{green}{white}{最大飞行速度}}大于\textbf{\fcolorbox{red}{white}{252}}的实体有几个？} & \makecell{How many entities have a \fcolorbox{green}{white}{\textbf{maximum}}\\ \fcolorbox{green}{white}{\textbf{flying speed}} greater than {\fcolorbox{red}{white}{\textbf{252}}}?}\\

            & \makecell[c]{top2} & \makecell[c]{\textbf{\fcolorbox{green}{white}{最大飞行速度}}等于\textbf{\fcolorbox{red}{white}{850}}的实体有几个？} & \makecell[c]{How many entities have a \fcolorbox{green}{white}{\textbf{maximum}}\\ \fcolorbox{green}{white}{\textbf{flying speed}} equal to {\fcolorbox{red}{white}{\textbf{850}}}?} \\
     
            & \makecell[c]{top3} & \makecell[c]{\textbf{\fcolorbox{green}{white}{最大飞行速度}}等于\textbf{\fcolorbox{red}{white}{745}}的实体有几个？} & \makecell[c]{How many entities have a \fcolorbox{green}{white}{\textbf{maximum}}\\ \fcolorbox{green}{white}{\textbf{flying speed}} equal to {\fcolorbox{red}{white}{\textbf{745}}}?} \\ [2ex]
     
            \Xhline{2pt}
        \end{tabular} 
    }
\end{table*}

Key information in the NLQs is highlighted using bold and boxes, with green boxes and red boxes denoting their presence and absence in the top NLQ, respectively.
It is evident that the demonstration examples selected by KI-Sim are more similar to the top NLQ.
This underscores the effectiveness of KI-Sim.

\section{Conclusion}

In this paper, we proposed a ChatGPT-based CQL generation framework, which consists of six components: 
an auxiliary model that predicted structural information for CQLs based on given NLQs, 
a proper noun matcher that extracted explicit proper nouns, 
a demonstration example selector that used KI-Sim to select demonstration examples, 
a prompt constructor that concatenated the NLQ, demonstration examples, and prior knowledge in the form of a Chain-of-Thought, 
a ChatGPT-based generation model that generated CQLs using the concatenated text, 
and an ensemble model that produced more reliable results by voting on diversified answers. 
Experimental results validate the effectiveness of our generative framework, achieving a remarkable second-place rank in the CCKS 2023 Question Answering with Knowledge Graph Inference for Unmanned Systems competition.

\subsubsection{Acknowledgements.} 
This work as supported by the Natural Science Foundation of China (62076133 and 62006117), and the Natural Science Foundation of Jiangsu Province for Young Scholars (BK20200463) and Distinguished Young Scholars (BK20200018).

\appendix

\section*{Appendix 1}
\label{appendix_prior_knowledge}
\subsection*{Original}
给定一个中文问题，用于对知识图谱的查询。问题既包含简单问题（实体属性、关系的单一查询），也包含复杂问题（显示约束、 隐式 约束、比较、布尔、多跳等）。请模仿提供的示例中体现的代码风格和语法，综合分析问题和实体名称、实体属性、实体标签、关系，生成一个符合cypher语法的查询语句，用于从知识图谱中获取答案。cypher语句应该包含以下部分：MATCH：用于匹配图数据库中的节点和 关 系，可以指定节点和关系的类型、属性和方向；WHERE：用于过滤匹配结果，可以使用逻辑运算符、比较运算符、字符串操作等；RETURN：用于返回查询结果，可以使用聚合函数、排序函数、限制函数等；WITH：用于将查询结果传递给下一个子句，可以使用聚合函 数、 排序函数、限制函数等。你需要识别问题中的实体、属性、标签、关系和约束条件，cypher语句用圆括号()表示节点，用方括号[]表示关系,用冒号:表示标签，用.表示属性，用箭头->表示关系方向。 例如，(n)，[r:`采用技术`]，[r:`厂商制造能力`]，n.`全重`，(n)-[r:`产国`]->(m)-[r1:`军种涉及项目`]->(l)。cypher语句用运算符如=, <, >, AND, OR, NOT来比较值和过滤结果。 用函 数如count(), min(), max(), avg(), sum()来对结果进行计算。 请注意，若问题为布尔型问题，cypher语句返回值是True或False 。例如,问题：MQ-4C是 2013年首飞吗？返回值：True。若问题为比较型问题，并询问实体，cypher语句返回值是实体名称。例如，问 题：M29式81毫米迫击 炮的口径和M1高射炮的相比，哪个更小？对应的cypher语句：MATCH (n) where n.name="M29式81毫米迫击炮" or n.name="M1高射炮" RETURN n.name ORDER BY n.`口径` asc limit 1返回值：M29式81毫米迫击炮若问题为比较实体属性值的问 题，cypher语句返回值是True 或False。例如，问题：扫描鹰无人机的交付数量是否比MQ-1"捕食者"无人机的多？对应的cypher语句 ：MATCH (n), (m) where n.name="扫描鹰无人机" and m.name="MQ-1"捕食者"无人机" RETURN n.`交付数量` > m.`交付数量`返回 值：False请自行判断问题类型，分析 问题中的逻辑关系，生成cypher语句。
\subsection*{Translated}
Given a Chinese question for querying a knowledge graph, the question includes simple queries about entity attributes and relationships, as well as complex queries involving explicit and implicit constraints, comparisons, boolean logic, and multi-hop scenarios. Emulating the provided code style and syntax, the analysis encompasses entities, attributes, labels, and relationships to generate Cypher Query Language (CQL) for knowledge graph answers.
CQL consists of several sections: MATCH, identifying nodes and relationships with specified types, properties, and directions; WHERE, filtering results with logical and comparison operators; RETURN, providing query outcomes using aggregation, sorting, and limiting functions; and WITH, forwarding results to subsequent clauses. 
You need to identify entities, attributes, labels, relationships, and constraint conditions in the question. In CQL syntax, use parentheses () to represent nodes, square brackets [] to represent relationships, colons : to represent tags, periods . to represent attributes, and arrows -> to indicate relationship directions. For example: [r:Technology Used], [r:Manufacturer Capability], n.Weight, (n)-[r:Origin Country]->(m)-[r1:Military Branch Involved Project]->(l).
Operators such as =, <, >, AND, OR, NOT are used to compare values and filter results. Functions such as count(), min(), max(), avg(), sum() are used to perform calculations on results. Note that if the question is boolean in nature, the CQL's return value is either True or False. For example, question: "Was MQ-4C first flown in 2013?" Return value: True.
If the question involves comparing entities and inquiring about an entity, the CQL's return value is the entity name. For example, question: "In comparison to the caliber of the M1 anti-aircraft gun, is the caliber of the M29 81mm mortar smaller?" Corresponding CQL: MATCH (n) where n.name="M29 81mm Mortar" or n.name="M1 Anti-Aircraft Gun" RETURN n.name ORDER BY n.`caliber` asc limit 1 Return value: M29 81mm Mortar.
If the question involves comparing entity attribute values, the CQL's return value is True or False. For example, question: "Is the delivery quantity of the ScanEagle UAV greater than that of the MQ-1 Predator UAV?" Corresponding CQL: MATCH (n), (m) where n.name="ScanEagle UAV" and m.name="MQ-1 Predator UAV" RETURN n.`delivery quantity` > m.`delivery quantity` Return value: False.
Please independently determine the question type, analyze logical relationships in the question, and generate CQL accordingly.

\section*{Appendix 2}
\label{appendix_cot_template}
\subsection*{Original}
已知问题原文为：“{question}”，让我们一步步思考：从原文中可以抽取出：实体：{entity}，标签：{tag}，属性：{attribute}，值：{value}，关系：{relation}，条件个数：{cn}，关系个数：{rn}，基于抽取结果，可得相应的cql语句为：{cql}
\subsection*{Translated}
Given the original question text: "\{question\}", let's think step by step: From the original text, we can extract: Entity: \{entity\}, Tag: \{tag\}, Attribute: \{attribute\}, Value: \{value\}, Relation: \{relation\}, Condition Count: \{cn\}, Relation Count: \{rn\}. Based on the extracted results, the corresponding CQL can be obtained as: \{cql\}

%
%
%
%

\bibliographystyle{unsrt}
\bibliography{reference}

\end{CJK*}

\end{document}